\documentclass[runningheads]{llncs}
\newif\ifblind
\blindfalse
\ifblind
    \usepackage[review,year=2026,ID=30]{eccv}
\else
    \usepackage{eccv}
    
\fi

\newif\ifarxiv
\arxivtrue

\usepackage{eccvabbrv}
\usepackage[T1]{fontenc}
\usepackage{graphicx}
\usepackage{booktabs}
\usepackage{amsmath}
\usepackage{amssymb}
\usepackage{microtype}
\usepackage[acronym]{glossaries}
\glsdisablehyper
\usepackage{tikz}
\usetikzlibrary{positioning,arrows.meta,fit}
\usepackage{ifthen}
\usepackage{tikz}
\usepackage{standalone}
\usepackage{comment}
\usepackage[accsupp]{axessibility} %
\usepackage{multirow}
\ifblind
    \usepackage[pagebackref,breaklinks,colorlinks,citecolor=eccvblue]{hyperref}
\else
    \usepackage[breaklinks,urlcolor=black,colorlinks,citecolor=eccvblue]{hyperref}
\fi

\usepackage{orcidlink}
\newacronymstyle{long-short-br}
{%
  \GlsUseAcrEntryDispStyle{long-short}%
}%
{%
  \GlsUseAcrStyleDefs{long-short}%
}
\setacronymstyle{long-short-br}
\newcommand*{\RL}[2][]{\textcolor{Rhodamine}{[\textbf{\ifthenelse{\equal{#1}{}}{RL}{RL(#1)}}: #2]}}

\usepackage{transparent}
\usepackage{tikz}
\ifarxiv
    \newcommand\copyrighttext{%
      \scriptsize Accepted to ABAW @ ECCV 2026. The final published version will be available on \textit{Springer}.}
    \newcommand\copyrightnotice{%
    \vspace{-5mm}
    \begin{tikzpicture}[remember picture,overlay]
    \node[anchor=south,yshift=80pt,xshift=0pt] at (current page.south) {\fbox{\transparent{0.85}\parbox{\dimexpr0.93\textwidth-\fboxsep-\fboxrule\relax}{\copyrighttext}}};
    \end{tikzpicture}%
    }
\else
\fi

\newcommand\major[1]{{#1}} %

\begin{document}
\title{Audio-Text Cross-Attention with\\Psycholinguistic Support Features for\\Ambivalence/Hesitancy Recognition}
\titlerunning{Audio-Text A/H Recognition with Psycholinguistic Features}
\author{Luiz F. B. F. Martins \and
Rodrigo W. Pisaia \and
Matheus M. Girardi \and\\
Isabella V.~Berkembrock \and
João A. Almeida \and
Andre G. Hochuli \and\\
Rayson Laroca \and
Alceu S. Britto Jr.}
\authorrunning{L. F. B. F. Martins et al.}
\institute{Pontifícia Universidade Católica do Paraná, Curitiba, Brazil\\
LIA - Artificial Intelligence Academic League\\
\email{ligaIA@ppgia.pucpr.br}}
\maketitle

\ifarxiv
    \copyrightnotice
\else
\fi

\newacronym{abaw}{ABAW}{Affective \& Behavior Analysis in-the-Wild}
\newacronym{ah}{A/H}{Ambivalence and Hesitancy}
\newacronym{ap}{AP}{Average Precision}
\newacronym{asr}{ASR}{Automatic Speech Recognition}
\newacronym{bah}{BAH}{Behavioral Ambivalence/Hesitancy}
\newacronym{mil}{MIL}{Multiple Instance Learning}

\ifblind
    \newcommand{\codeURL}{[\textit{hidden for review}]}
    
\else
    \newcommand{\codeURL}{\url{https://github.com/Liga-de-IA-PUCPR/abaw-11-ah-challenge/}}
\fi

\begin{abstract}
We present a frame-independent audio--text system for the 3rd Ambivalence/Hesitancy Video Recognition Challenge at the 11th \gls*{abaw} Workshop.
Videos are divided into overlapping 5-s windows aligned with transcript timestamps. Each window combines prosodic audio descriptors, emotion-oriented RoBERTa embeddings, and 74 psycholinguistic features representing uncertainty, hedging, and attitudinal conflict. Temporal cross-attention fuses audio and text, while the support features condition gated \gls*{mil} pooling. A five-seed ensemble achieves an average precision of 0.875 and a macro-F1 of 0.722 on the 525-video labeled public-test split. 
Notably, our submission ranked third overall on the official challenge leaderboard, with a macro-F1 of 0.7455.
Source code is available at~\codeURL.

\end{abstract}

\section{Introduction}
\label{sec:introduction}
\glsresetall

\gls*{ah} are relevant behavioral signals in digital behavior-change
interventions because they may indicate that a person is uncertain about, or
simultaneously attracted to and resistant toward, a proposed
action~\cite{delatorre2024heartsteps}. Ambivalence is not equivalent to
neutrality or simple indecision: it arises when positive and negative
evaluations of the same object are simultaneously accessible, producing
evaluative conflict even when their net difference is close to
zero~\cite{kaplan1972,thompson1995,schneider2017mixed}. Such conflict can
increase information processing and weaken attitude--intention consistency,
making ambivalence especially relevant when a person is considering
behavioral change~\cite{jonas1997effects,conner2021}. Hesitancy is the
temporally unfolding expression of reduced commitment and may be conveyed by
delayed responses, silent or filled pauses, lengthening, hedges, repairs,
slower speech, and rising or more variable
intonation~\cite{smith1993,brennan1995,swerts2005,jiang2017}.
As these markers reflect partially distinct linguistic and acoustic
processes, hesitancy should not be reduced to the occurrence of a single
filler or pause~\cite{betz2023cognitive}.

Recognizing \gls*{ah} in videos is challenging for two main reasons. First, the
relevant evidence may occur only during a short portion of a video rather
than persist throughout the complete response. A positive video may contain
long neutral or fluent stretches interleaved with a brief pause, hedge,
self-correction, or explicit evaluative conflict. Consequently, the
video-level label does not imply that every temporal segment contains
diagnostic evidence. This setting is related to weakly supervised temporal
localization, in which a global label may be explained by a sparse subset of
discriminative snippets whose relevance must be inferred without using dense
annotations during training~\cite{ren2023proposal}.
Uniform temporal averaging can therefore dilute localized \gls*{ah} cues
with substantially longer non-diagnostic intervals. Second, useful evidence
is distributed across modalities. Response latency, pauses, speech timing,
pitch variation, epistemic hedges, contrastive constructions, and
conflicting positive and negative evaluations may all contribute to the
final decision~\cite{qi2025mfgcn,shang2024multimodalfusion}. Explicitly
modeling these cues is especially relevant when the amount of labeled data
is limited and generic high-dimensional embeddings may not reliably expose
the constructs of interest.

Such challenges motivate our central research question: can
attention-based temporal aggregation combined with explicit prosodic and
psycholinguistic support features improve audio--text \gls*{ah} recognition
over generic embeddings and uniform temporal pooling, without relying on
visual-frame input? Our objective is to develop a frame-independent
audio--text method for video-level \gls*{ah} recognition that identifies
informative temporal segments while explicitly representing behavioral
markers of uncertainty and attitudinal conflict. Although visual behavior
may provide complementary evidence, audio and language already contain
well-established indicators of \gls*{ah}; studying these modalities without
visual frames also makes it possible to isolate their contribution and
removes the need for frame decoding and visual feature extraction.

\major{We hypothesize that~\textbf{(H1)} attention-based \gls*{mil} pooling provides a more effective video-level aggregation mechanism than uniform mean pooling, as A/H evidence is temporally sparse in the BAH corpus.}
We further hypothesize that \textbf{(H2)}~explicit support features encoding prosodic
uncertainty, textual hedging, and attitudinal conflict provide a
complementary inductive bias beyond what generic audio and text embeddings
capture alone.

To investigate these hypotheses, we develop an audio--text architecture that
operates exclusively on audio and timestamped transcripts while preserving
the video-level prediction setting.
Videos are divided into overlapping 5-s
windows that match the typical duration of an \gls*{ah} episode
in the \gls*{bah} dataset~\cite{bahdataset}.
For every window, we extract prosodic audio descriptors, an
emotion-oriented text embedding, and an interpretable support vector
grounded in psycholinguistic descriptions of uncertainty and ambivalence.
Cross-attention combines the aligned audio and text representations, after
which the support features are incorporated into the window representation.
Gated \gls*{mil} pooling then assigns greater weight to the most informative
windows, and a five-seed ensemble is used to reduce training~variance.

The main contributions of this work are threefold:
\begin{itemize}
    \item A frame-independent audio--text architecture that combines temporal
    cross-attention with gated \gls*{mil} pooling for localized video-level
    \gls*{ah} recognition;

    \item A 74-dimensional psycholinguistic support representation encoding
    question context, prosodic uncertainty, textual hedging, emotion
    dynamics, and attitudinal ambivalence;

    \item An experimental analysis of temporal aggregation and explicit
    support features, showing that the support representation provides the
    largest observed improvement.
\end{itemize}

The experimental results on a 525-video labeled public test set show that replacing uniform mean pooling
with attention-based \gls*{mil} increases average precision from 0.828 to
0.835 and macro-F1 from 0.701 to 0.705. Adding the complete support
representation and averaging the predictions of five independently
initialized models further increases average precision to 0.875 and
macro-F1 to 0.722. These results indicate that explicit prosodic and
psycholinguistic descriptors provide useful complementary information for
audio--text \gls*{ah} recognition, while the smaller improvement obtained
from \gls*{mil} provides more limited evidence for the advantage of learned
temporal aggregation.

The remainder of this paper is organized as follows.
\Cref{sec:related} reviews the literature on prosodic and linguistic uncertainty, attitudinal ambivalence, multimodal fusion, and temporal aggregation.
\Cref{sec:proposed} presents the architecture and
feature representations. \Cref{sec:protocol} describes the dataset, training
procedure, and evaluation protocol. \Cref{sec:results} reports and discusses
the results. Finally, \Cref{sec:conclusion} summarizes the findings and
outlines directions for future~work.

\section{Related Work} \label{sec:related} 

Research on \gls*{ah} recognition draws on speech prosody, linguistic uncertainty, attitudinal ambivalence, and multimodal temporal aggregation. These complementary areas motivate the domain-informed support features, cross-modal fusion, and temporal pooling used in our approach. 

\subsection{Prosodic and linguistic markers of uncertainty} 
Speech conveys a speaker's degree of certainty through timing, intonation, loudness, and voice quality. Doubtful utterances typically exhibit longer response latencies, more pauses, slower speech, rising terminal intonation, lower or less stable intensity, and greater fundamental-frequency variability~\cite{brennan1995,smith1993,swerts2005,jiang2017}. Such features can be operationalized through automatic estimates of speech and articulation rate based on syllable-nucleus detection \cite{dejong2009,cucchiarini2000}, together with compact voice-quality descriptors such as jitter and shimmer~\cite{eyben2016}. 

The literature distinguishes uncertainty, as a speaker state, from disfluencies, as discrete, observable events. Filled pauses can be detected from acoustic and textual representations, with acoustic models generally providing stronger frame-level evidence and multimodal combinations yielding additional gains~\cite{chatziagapi2022filledpause}. However, filler counts alone do not represent continuous variations in prosodic uncertainty and may be unreliable when derived from \gls*{asr} transcripts. We therefore model timing, pitch, loudness, speech rate, and voice quality directly rather than treating filled-pause detection as the primary task. Text provides complementary evidence through hedges and other epistemic~markers. 

The CoNLL-2010 shared task~\cite{farkas2010} established uncertainty detection as a context-dependent problem: expressions such as ``may,'' ``suggest,'' and ``I think'' function as hedges only in particular linguistic contexts.
Hedge detection has also been approached through distant supervision and shallow linguistic features, which are especially useful in label-scarce settings~\cite{ganter2009}. These findings motivate explicit features for graded epistemic commitment, contrast, and polarity shifts.

\subsection{Attitudinal ambivalence and mixed affect} 

Psychological models describe ambivalence as the simultaneous activation of positive and negative, or approach and avoidance, tendencies. Behavioral studies have shown that conflicting appetitive and aversive evidence affects gaze and response trajectories, producing measurable approach--avoidance conflict \cite{chen2025approach,garciaguerrero2023action}.

Psychometric models therefore represent positive and negative components separately. Kaplan's ambivalence-indifference problem shows that a net polarity score cannot distinguish weak positive and negative evidence from strong but conflicting evidence~\cite{kaplan1972}. Conflict can instead be represented using quantities such as $\min(P,N)$ and the similarity-intensity index using \cref{eq:sim}.

\begin{equation}
\label{eq:sim}
\frac{P+N}{2} - |P-N| \; .
\end{equation}

The index increases when positive and negative components are simultaneously strong~\cite{thompson1995}.
Related work also distinguishes cognitive--affective inconsistency from subjectively experienced ambivalence~\cite{conner2021}.
A similar principle appears in multi-label emotion recognition, where opposite-valence emotions may co-occur rather than being treated as mutually exclusive.
Models such as MEDA explicitly capture correlations among emotions and can therefore represent mixed affective states~\cite{deng2020}.
Together, these findings motivate separate positive and negative intensities, ambivalence indices, emotion-distribution uncertainty, and temporal variation in valence.

\subsection{Cross-modal fusion and temporal aggregation} 

Attention mechanisms provide a flexible framework for modeling interactions between modalities~\cite{vaswani2017}. In affective computing, audio-text cross-attention has been used to capture complementary semantic and paralinguistic information, thereby improving emotion recognition over unimodal or simple late-fusion systems \cite{sun2021mcsan}. 

Related work has extended cross-modal attention to visual streams \cite{das2024avater}. These studies motivate our use of audio-to-text cross-attention over aligned temporal windows.
As \gls*{ah} labels are provided at the video level while evidence may occur only in a few temporal segments, \acrfull*{mil} provides a natural formulation. A video can be represented as a bag of windows, with attention-based pooling learning the contribution of each instance \cite{ilse2018}.

Similar formulations have been applied to temporally localized psychological and clinical evidence, including stress, autism spectrum disorder, and pain recognition \cite{brugge2025stressmil,duan2024facialinstance}. Hard-top-$k$ aggregation may produce incomplete localization when the evidence is discontinuous, motivating softer attention-based alternatives \cite{wang2026semanticprototype}. We therefore use gated soft \gls*{mil} pooling~\cite{ilse2018}, allowing multiple windows to contribute with learned continuous weights rather than selecting a fixed number of instances. 

\subsection{A/H recognition systems}

The \gls*{bah} dataset~\cite{bahdataset} and its associated challenge are recent, and most directly related systems are challenge submissions or preprints evaluated on different validation, public-test, or private-test splits.
Their reported scores should therefore be compared cautiously.
Most existing systems combine visual, audio, and textual~information.

\major{The broader evolution of the ABAW benchmark and its associated affective and behavioral analysis tasks has been documented in successive workshop and competition reports~\cite{kollias2025advancements,kollias2026from}.}

Cabacas-Maso et al.~\cite{cabacasmaso2026calibrated} ensemble fusion models over face, audio, text, and pose representations. Bui et al.~\cite{bui2026cfnet} use visual, audio, and textual embeddings with speaker-normalized features and a certainty-weighted loss. Bakin and Savchenko~\cite{bakin2026hsemotion11th} combine frame-level face, audio, and text classifiers through late fusion and transcript-based gating, while Kumar et al.~\cite{kumar2026simple} enrich multimodal embeddings with hand-crafted hesitation markers. Ryumina et al.~\cite{ryumina2026teamras} instead use text as an anchor modality and introduce gated residual contributions from audio, face, and scene streams.

Related baselines for earlier or differently evaluated versions of the task include video architectures and multimodal-LLM prompting~\cite{gonzalez2026personalized}.
In contrast to these predominantly audiovisual systems, our method uses only audio and timestamped transcripts.
It compensates for the absence of a visual stream through explicit prosodic, epistemic, and attitudinal support features.
These descriptors are fused with the cross-attended audio--text representation before gated \gls*{mil} pooling, allowing them to influence both window representations and temporal aggregation weights.

\section{Proposed Method}
\label{sec:proposed}

\begin{figure*}[t]
    \centering
    \includegraphics[width=\textwidth]{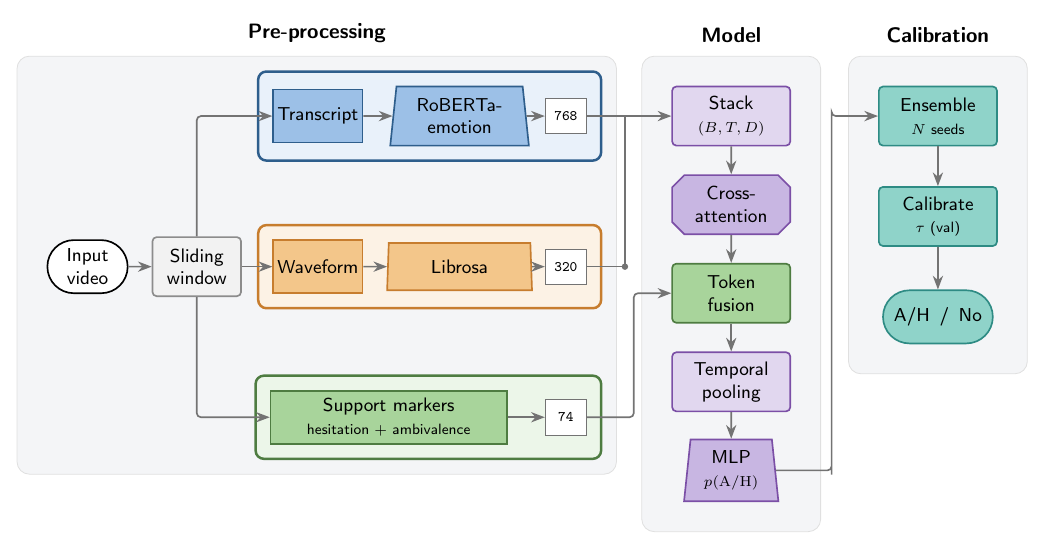}
    
    \vspace{-1mm}
    
    \caption{Overview of the audio--text cross-attention architecture with psycholinguistic support features.}
    \label{fig:arch}
\end{figure*}

\Cref{fig:arch} provides an overview of the proposed architecture, which integrates audio-text cross-attention, psycholinguistic support features, gated \gls*{mil} pooling, and a video-level classifier.

\subsection{Pre-processing and temporal alignment}
The audio track is extracted from each video, converted to mono 16-kHz FLAC, and segmented into 5-s windows with a 2.5-s hop.
The transcripts in the \gls*{bah} dataset include Whisper-generated text and timestamps.
We use these timestamps to concatenate all transcript segments that overlap each audio window.
Timestamp resets at internal Whisper segment boundaries are corrected by accumulating an offset, which produces a monotonic timeline.

The target is the global video label $y_v\in\{0,1\}$.
The released participant-wise splits prevent the same participant from appearing in more than one partition.
The complete video is represented as a variable-length sequence of $T$ aligned~windows.

\subsection{Window representations}
\label{sec:window-representations}

Each temporal window $t$ is represented by three complementary vectors:
an audio descriptor $\mathbf{a}_t\in\mathbb{R}^{320}$, a text embedding
$\mathbf{x}_t\in\mathbb{R}^{768}$, and a psycholinguistic support vector
$\mathbf{s}_t\in\mathbb{R}^{74}$. The audio and text vectors provide generic
acoustic, semantic, and affective representations, whereas the support vector
explicitly encodes cues associated with attitudinal ambivalence and
hesitancy.

\paragraph{Audio representation.}
The 320-dimensional audio descriptor is extracted with \texttt{librosa} and
contains MFCCs and their first- and second-order derivatives, spectral
centroid, bandwidth, roll-off, flatness, spectral contrast, chroma,
zero-crossing rate, RMS energy, pitch statistics, voiced fraction, and tempo.
When applicable, frame-level descriptors are summarized by their mean,
standard deviation, minimum, and maximum. This representation was selected
after preliminary experiments in which generic self-supervised speech
embeddings did not improve~performance.

\paragraph{Text representation.}
Text is encoded with the TweetEval RoBERTa emotion
encoder\footnote{\url{https://huggingface.co/cardiffnlp/twitter-roberta-base-emotion}}
\cite{tweeteval,roberta}. We remove the classification head, apply
attention-masked mean pooling over the token representations, and perform
$\ell_2$ normalization. The resulting 768-dimensional vector represents the
semantic and affective content of each window.

\paragraph{Psycholinguistic support representation.}
The support vector comprises three blocks: a 7-dimensional question-context
descriptor, an 18-dimensional acoustic--prosodic block, and a 49-dimensional
textual block. \Cref{tab:support} summarizes their composition, dimensionality,
and temporal granularity.

\begin{table}[htb!]
\caption{Composition and temporal granularity of the psycholinguistic support
vector. Response-level features are broadcast to every window.}
\label{tab:support}
\centering
\begin{tabular}{lllr}
\toprule
Block & Feature group & Granularity & Dim. \\
\midrule
Context
    & Question type
    & Response
    & 7 \\
\midrule
\multirow{5}{*}{Acoustic--prosodic}
    & Timing and pauses
    & Window
    & 6 \\
    & Speech rate and articulation
    & Window
    & 3 \\
    & Pitch and intonation
    & Window
    & 4 \\
    & Loudness
    & Window
    & 2 \\
    & Voice quality and uncertainty score
    & Window
    & 3 \\
\cmidrule(lr){2-4}
    & \multicolumn{2}{l}{Acoustic--prosodic subtotal}
    & 18 \\
\midrule
\multirow{4}{*}{Textual}
    & Ambivalence indices
    & Response
    & 20 \\
    & Emotion dynamics
    & Mixed
    & 6 \\
    & Contrast and polarity shift
    & Window
    & 5 \\
    & Epistemic hedges
    & Window
    & 18 \\
\cmidrule(lr){2-4}
    & \multicolumn{2}{l}{Textual subtotal}
    & 49 \\
\midrule
\multicolumn{3}{l}{Total}
    & 74 \\
\bottomrule
\end{tabular}
\end{table}

The acoustic support features are computed from the raw waveform using
\texttt{librosa} for timing, pause, and loudness descriptors and Praat through
Parselmouth for F0, jitter, and shimmer. The analysis uses a frame length of
2048 samples and a hop of 512 samples at 16\,kHz. Textual features are computed
from the \gls*{asr} transcript using the VADER sentiment lexicon~\cite{hutto2014} and
the four-class output of the TweetEval emotion model~\cite{tweeteval}, with
labels \{anger, joy, optimism,~sadness\}.

\paragraph{Feature granularity.}
Each video contains a single question--answer pair, with approximately 24
words per window and 130 words per complete response, distributed over about
nine windows. The question type is constant throughout the response and is
therefore represented by a 7-dimensional one-hot vector broadcast to every
window.

Ambivalence is also modeled primarily at the response level because opposing
evaluations may occur in different windows. A single short window may therefore
contain insufficient sentiment-bearing content to estimate positive and
negative evidence reliably. We compute the 20 ambivalence features over the
reconstructed transcript and broadcast them to every window. Four emotion
dynamics are likewise computed at the response level: the standard deviation
and range of window-level valence, pole conflict, and overall response valence.

All acoustic--prosodic features are computed locally for each window. The
epistemic-hedge and contrast/polarity-shift groups are also window-specific.
The remaining two emotion-dynamics features---emotion entropy and the gap
between the two highest emotion probabilities---are computed per window.
Thus, the emotion-dynamics block has mixed granularity, as indicated in
\Cref{tab:support}.

\paragraph{Acoustic--prosodic features.}
The 18 acoustic--prosodic features quantify the speaker's state of
(un)certainty rather than detecting isolated disfluency events such as filled
pauses~\cite{jiang2017,brennan1995}. A voice-activity detector based on
\texttt{librosa.effects.split} with $\texttt{top\_db}=30$ segments each window
into speech intervals and internal silences.

\begin{itemize}
\item \emph{Timing and pauses} ($6$) comprise onset latency, the proportion of
low-energy frames, pauses per second, mean pause duration, the proportion of
time spent in long pauses, and voiced fraction. Low-energy frames have RMS
below $0.1$ times the window peak; pauses and long pauses correspond to
silences of at least $0.15$\,s and more than $0.5$\,s,
respectively~\cite{brennan1995,smith1993,swerts2005,jiang2017};

\item \emph{Speech rate and articulation} ($3$) comprise speech rate, articulation
rate, and phonation ratio. Syllable nuclei are detected as intensity peaks
within 25\,dB of the loudest frame and separated by a dip of at least 2\,dB,
following de~Jong and Wempe~\cite{dejong2009}. The measures are derived from
the number of nuclei, total duration, and phonation
time~\cite{cucchiarini2000,jiang2017};

\item \emph{Pitch and intonation} ($4$) are computed from the Praat F0 contour over
voiced frames using autocorrelation in the 65--500\,Hz range. They comprise
the coefficient of variation, semitone range, global slope, and terminal slope
over the final 30\% of voiced frames. Greater variability and rising terminal
intonation are associated with uncertainty~\cite{jiang2017,brennan1995};

\item \emph{Loudness} ($2$) is represented by the mean and coefficient of variation
of frame-level RMS energy, since lower and less stable intensity may indicate
doubt~\cite{jiang2017};

\item \emph{Voice quality and uncertainty score} ($3$) comprise local jitter and
shimmer, computed with Praat and motivated by the eGeMAPS parameter
set~\cite{eyben2016}, and a fixed-weight uncertainty score used only for
inspection.\footnote{
$s=\mathrm{clip}\!\left(
0.20\,s_{\mathrm{lat}}
+0.20\,s_{\mathrm{pause}}
+0.15\,s_{\mathrm{rise}}
+0.15\,s_{\mathrm{F0cv}}
+0.15\,s_{\mathrm{loud}}
+0.15\,s_{\mathrm{slow}},
0,1\right)$.
Each component is clipped to $[0,1]$, and the F0-related terms are gated by
the voiced fraction.}

\end{itemize}

\paragraph{Textual features.}
The 49 textual features represent ambivalence, emotion dynamics, discourse
contrast, polarity shifts, and epistemic hedging.

\begin{itemize}
\item \emph{Ambivalence indices} ($20$) are computed from two polarity sources:
VADER positive and negative proportions and the emotion-model probabilities,
where positive evidence is defined as joy plus optimism and negative evidence
as anger plus sadness. For each source, we retain positive intensity $P$,
negative intensity $N$, and four psychometric measures:
\begin{equation}
\frac{P+N}{2},
\qquad
|P-N|,
\qquad
\min(P,N),
\qquad
\frac{P+N}{2}-|P-N|.
\label{eq:ambivalence-indices}
\end{equation}
These correspond to mean intensity, polar discrepancy, Kaplan's
minimum~\cite{kaplan1972}, and the Griffin similarity--intensity
index~\cite{thompson1995,conner2021}. Retaining $P$ and $N$ separately avoids
mapping neutral and ambivalent responses to the same net score
$P-N$~\cite{kaplan1972}.

The remaining eight ambivalence features describe an approach--avoidance axis.
Desire and resistance markers are identified using hand-built lexicons released
with the code. Their frequencies are represented as raw counts, per-word rates,
and per-100-word rates, together with the product and minimum of the desire and
resistance components.

\item \emph{Emotion dynamics} ($6$) include two window-level and four response-level
measures. Window-level features are the Shannon entropy of the four-class
emotion distribution and the gap between its two highest probabilities.
Response-level features are the standard deviation and range of window
valence $P-N$, pole conflict $\min(P,N)$, and overall response
valence~\cite{deng2020}.

\item \emph{Contrast and polarity shift} ($5$) comprise the frequency of
contrastive discourse markers in raw, per-word, and per-100-word forms, the
absolute difference between VADER compound scores in the first and second
halves of the window, and a binary indicator of whether polarity changes sign.
Contrastive markers include \emph{but}, \emph{however}, \emph{although},
\emph{though}, \emph{yet}, \emph{on the other hand}, and
\emph{at the same time}.

\item \emph{Epistemic hedges} ($18$) represent six categories: first-person
epistemic expressions, adverbial hedges, approximators, explicit uncertainty,
agentless attribution, and the aggregate hedge count. Each category is
represented as a raw count, a per-word rate, and a per-100-word
rate~\cite{farkas2010,ganter2009}.
\end{itemize}

\subsection{Cross-attention and token fusion}
Audio and text are independently projected to a common width $d=512$.
Multi-head cross-attention uses the audio sequence as query and the aligned text sequence as key and value:
\begin{equation}
\mathbf{H}=\mathrm{LN}\!\left(\mathbf{A}+\mathrm{MHA}(\mathbf{A},\mathbf{X},\mathbf{X})\right),
\end{equation}
where padding windows are masked and LN denotes layer normalization.

The heterogeneous support features are normalized over valid windows only, projected to 512 dimensions, and concatenated with the cross-modal representation before another linear projection:
\begin{equation}
\widetilde{\mathbf{h}}_t=\mathrm{LN}\left(W_f[\mathbf{h}_t\Vert g(\mathbf{s}_t)]\right).
\end{equation}
This token fusion places the interpretable cues before temporal aggregation.
Consequently, a window containing strong uncertainty or conflict cues can receive greater \gls*{mil} attention.
Experiments showed that this placement was more effective than concatenating the support vector only after video pooling.

\subsection{\gls*{mil} pooling, classifier, and ensemble}
Gated attention pooling assigns a score to each valid window~\cite{ilse2018}:
\begin{align}
q_t &= \mathbf{w}^{\top}\!\left[\tanh(V\widetilde{\mathbf{h}}_t)\odot\sigma(U\widetilde{\mathbf{h}}_t)\right],\\
\alpha_t &= \frac{\exp(q_t)}{\sum_{j=1}^{T}\exp(q_j)}, \qquad
\mathbf{z}=\sum_{t=1}^{T}\alpha_t\widetilde{\mathbf{h}}_t.
\end{align}
The video vector $\mathbf{z}$ passes through a 512-unit ReLU layer, dropout (0.3), and a binary output layer.
The model is optimized with binary cross-entropy and AdamW using a learning rate of $3\times10^{-4}$, weight decay of 0.05, a batch size of 32, gradient clipping at 1.0, and early stopping based on validation average precision.

Five models with seeds $\{42,1,2,3,4\}$ are trained with identical settings, and
their video probabilities are averaged into a single score $s_v\in[0,1]$ per video.

\subsection{Threshold calibration}

Binary predictions are obtained by applying a threshold~$\tau$ to the
ensemble score~$s_v$. The threshold is selected exclusively on the
validation split and frozen for test evaluation. As the validation
set contains only 124 videos, directly maximizing macro-F1 over
$\{0,0.01,\ldots,1\}$ may be unstable.

For the ensemble, we smooth the validation macro-F1 curve with a moving
average of width 0.1 and define the near-optimal plateau, as denoted in \cref{eq:tr}.

\begin{equation}
\label{eq:tr}
\mathcal{P}=\bigl\{\tau : \widetilde{F}_1(\tau)\ge\textstyle\max_{\tau'}\widetilde{F}_1(\tau')-0.01\bigr\}\;.
\end{equation}

\noindent We select the center of its longest contiguous interval. For individual
models, we additionally evaluate base-rate matching, setting $\tau$ to the
$(1-p_+)$ quantile of validation scores using \cref{eq:ct}. Both strategies use validation information only.

\begin{equation}
\label{eq:ct}
p_+=\frac{1}{|\mathcal{V}|}\sum_{v\in\mathcal{V}} y_v,
\qquad
\tau=Q_{1-p_+}\bigl(\{s_v\}_{v\in\mathcal{V}}\bigr).
\end{equation}

\section{Experimental Protocol}
\label{sec:protocol}
The \gls*{bah} dataset~\cite{bahdataset} contains 1,427 labeled videos from 300 Canadian participants.
The released splits contain 778 training, 124 validation, and 525 public-test videos.
Our final configuration fits the model weights on the training set, while the threshold used for the reported macro-F1 is calibrated on the validation set.

We evaluate video-level recognition using macro-F1 across the positive and negative classes and additionally report positive-class \gls*{ap}.
All reported results were produced by the supplied implementation on the labeled public test split.
All experiments were conducted on a single Apple M3 Pro chip~(14-core GPU, 18\,GB unified memory), using the Metal Performance Shaders~(MPS) backend for GPU acceleration.

\section{Results and Discussion}
\label{sec:results}

\Cref{tab:results} reports the main results on the 525-video labeled public-test split of \gls*{bah}. The experiments isolate the effects of temporal pooling, the acoustic and textual support-feature groups, the complete support representation, and probability averaging across independently initialized models.

\begin{table}[htb!]
\caption{Performance on the BAH public test split containing 525 labeled videos~\cite{bahdataset}.}
\label{tab:results}
\centering
\begin{tabular}{lcc}
\toprule
Method & Macro-F1 & AP \\
\midrule
\multicolumn{3}{l}{\textit{Comparison methods}} \\
\hspace{3mm}ConflictAwareAH~\cite{bekhouche2026conflict} (Fennec) & 0.694 & N/A \\
\hspace{3mm}Mean-pooling baseline & 0.701 & 0.828 \\
\addlinespace
\multicolumn{3}{l}{\textit{Proposed MIL variants}} \\
\hspace{3mm}Audio + text, without support features & 0.705 & 0.835 \\
\hspace{3mm}Audio support only & 0.636 & 0.744 \\
\hspace{3mm}Text support only & 0.698 & 0.842 \\
\hspace{3mm}All support features, single seed & 0.710 & 0.857 \\
\hspace{3mm}All support features, five-seed ensemble & \textbf{0.722} & \textbf{0.875} \\
\bottomrule
\end{tabular}
\end{table}

\paragraph{Temporal aggregation.} Replacing uniform temporal mean pooling with attention-based \gls*{mil}, while retaining only the audio and text embeddings, increases \gls*{ap} from 0.828 to 0.835 and macro-F1 from 0.701 to 0.705. The improvement is consistent across both metrics but small in magnitude.
\major{This result is directionally consistent with H1, as learned temporal weighting yields a modest improvement over uniform averaging. However, the gain was obtained from a single training run and is comparable in magnitude to the seed-to-seed variation observed elsewhere in our experiments. We therefore regard this result as suggestive rather than conclusive evidence in support of H1. Furthermore, the learned \gls*{mil} weights are neither visualized nor compared with the corpus' time-resolved \gls*{ah} annotations in the present study, although such annotations make this analysis possible. We leave this investigation to future work and, accordingly, interpret the learned weights only as internal aggregation coefficients rather than validated indicators of the temporal location of \gls*{ah} evidence. Our empirical conclusions are therefore restricted to video-level~recognition.}

\paragraph{Acoustic support features.}
Adding the acoustic--prosodic support block decreases macro-F1 from 0.705 to 0.636 and \gls*{ap} from 0.835 to 0.744.
These features may be noisy in short or weakly voiced windows and partly redundant with the 320-dimensional base audio representation, which already includes timing, pitch, energy, and spectral descriptors.
The acoustic support block therefore provides no independent benefit in the present~configuration.

\paragraph{Textual support features.}
Textual support increases AP from 0.835 to 0.842 but decreases macro-F1
from 0.705 to 0.698. The explicit hedging, polarity, emotion, and
ambivalence markers therefore improve ranking quality, although this gain
does not transfer to threshold-dependent classification, possibly because
of calibration sensitivity or overlapping class-score distributions.

\paragraph{Complete support representation.}
Combining contextual, acoustic, and textual support produces the strongest single-model result, increasing \gls*{ap} from 0.835 to 0.857 and macro-F1 from 0.705 to 0.710.
This controlled comparison supports H2 for the complete representation rather than for each feature group independently.
The full representation outperforms both isolated blocks, suggesting complementarity among the support cues, although the current ablation does not directly test their interactions.
As the support features are fused before temporal pooling, they may influence both the window representations and the internal \gls*{mil} weighting. Establishing whether these learned weights correspond to meaningful \gls*{ah} episodes, however, requires a dedicated temporal analysis and is therefore beyond the video-level evaluation performed here.

\paragraph{Ensemble effect.}
Averaging five full-support models increases AP from~0.857 to~0.875 and
macro-F1 from~0.710 to~0.722.
As shown in \Cref{tab:ensemble}, most of the
gain is already obtained with three models, and performance saturates
thereafter. Five seeds yield the highest AP and stable macro-F1 under
smooth calibration, although they do not dominate every thresholding
strategy. The ensemble gain is therefore distinct from, and additional to,
the contribution of the support~representation.

\begin{table}[htb!]
\centering
\caption{Effect of ensemble size on the test set under smooth and base-rate threshold calibration, with $\star$ indicating the selected configuration.}
\setlength{\tabcolsep}{5pt}
\label{tab:ensemble}
\small
\begin{tabular}{lccc}
\toprule
Seeds & F1$_\text{test}$ (smooth) & F1$_\text{test}$ (base rate) & AP$_\text{test}$ \\
\midrule
1 (single)      & 0.7107 & 0.7081 & 0.8572 \\
3               & 0.7213 & 0.7252 & 0.8719 \\
5 $\star$       & 0.7226 & 0.7191 & \textbf{0.8750} \\
7               & 0.7245 & 0.7124 & 0.8726 \\
\bottomrule
\end{tabular}
\end{table}

\paragraph{\major{Official challenge comparison.}}

\Cref{tab:challenge} reports the final leaderboard of the 3rd Ambivalence/Hesitancy (AH) Video Recognition Challenge. Our submission~(PUCPR) achieved a macro-F1 of 0.7455 and ranked third overall, behind AffectVision~(0.7959) and RAS~(0.7824), while narrowly outperforming AIM for Emo~(0.7431). The winning AffectVision approach combines affect-specialized textual, acoustic, and visual representations with interpretable hesitation cues through Affective Marker Fusion, followed by an AP-weighted ensemble with a fixed decision threshold~\cite{kumar2026simple}. The second-ranked RAS approach adopts a text-centered multimodal strategy in which linguistic information serves as the anchor and acoustic, facial, and scene representations provide complementary information through gated residual adjustments~\cite{ryumina2026teamras}. In contrast, our system relies exclusively on audio and timestamped transcripts, without processing visual frames. Its third-place result therefore demonstrates competitive challenge performance while retaining a frame-independent design. The leaderboard score is reported separately from the controlled experiments in \Cref{tab:results}, which use the labeled public-test split to analyze the individual components of the proposed method.

\begin{table}[htbp]
\centering
\caption{\major{Final leaderboard of the 3rd Ambivalence/Hesitancy (AH) Video Recognition Challenge. Our submission~(PUCPR) ranked third overall.}}
\label{tab:challenge}
\small
\setlength{\tabcolsep}{4pt}

\begin{tabular}{@{}clc@{}}
\toprule
\major{Rank} & \major{Team} & \major{Macro-F1} \\
\midrule
\major{1} & \major{AffectVision} & \major{0.7959} \\
\major{2} & \major{RAS} & \major{0.7824} \\
\major{\textbf{3}} & \major{\textbf{PUCPR (Ours)}} & \major{\textbf{0.7455}} \\
\major{4} & \major{AIM for Emo} & \major{0.7431} \\
\major{5} & \major{AIWELL} & \major{0.7367} \\
\major{5} & \major{CASIA-26} & \major{0.7367} \\
\major{6} & \major{Tamimi} & \major{0.7364} \\
\bottomrule
\end{tabular}
\quad
\begin{tabular}{@{}clc@{}}
\toprule
\major{Rank} & \major{Team} & \major{Macro-F1} \\
\midrule
\major{7} & \major{CPR} & \major{0.7167} \\
\major{8} & \major{IUSD} & \major{0.6841} \\
\major{9} & \major{MINDH Lab} & \major{0.6324} \\
\major{10} & \major{HSEmotion} & \major{0.5623} \\
\major{11} & \major{GGL} & \major{0.5136} \\
\major{12} & \major{AVULAB} & \major{0.4858} \\
\major{--} & \major{Baseline} & \major{0.3448} \\
\bottomrule
\end{tabular}

\end{table}

\paragraph{Overall interpretation.}
The results provide limited but directionally positive evidence for H1 and
stronger evidence for H2. MIL yields a small improvement over mean pooling,
whereas the complete support representation provides the clearest
single-model gain. Acoustic and textual support behave differently in
isolation, and probability averaging provides a further improvement,
particularly in AP. All results are obtained without visual-frame features,
establishing a frame-independent audio--text baseline for video-level A/H
recognition.

\section{Conclusions}
\label{sec:conclusion}

\glsreset{ah}
\glsreset{mil}

We presented a frame-independent audio--text approach for video-level
\gls*{ah} recognition. The method combines prosodic audio
descriptors and emotion-oriented text embeddings through cross-attention,
incorporates psycholinguistic support features before gated \gls*{mil}
pooling, and averages five independently initialized models.

The experiments separate the effects of temporal aggregation, support
features, and ensembling. Gated \gls*{mil} provides a small improvement over
mean pooling, offering limited support for H1. The complete support
representation yields the strongest single-model result and supports H2,
although its acoustic and textual components are not consistently beneficial
when used independently. The five-seed ensemble further improves performance,
reaching an \gls*{ap} of 0.875 and a macro-F1 of 0.722. These results show
that explicit uncertainty and ambivalence cues complement generic audio and
text representations without requiring visual-frame~processing.
Notably, on the official challenge leaderboard, our submission achieved a macro-F1 of 0.7455 and ranked third overall.

The study is limited by the small, single-dataset evaluation, possible
instability in threshold calibration, dependence on Whisper-generated
transcripts, and the poor performance of the acoustic support block when used
alone. Excluding visual information also leaves potentially complementary
facial and gestural evidence~unexplored.

Future work should evaluate the method on additional datasets, characterize the effect of \gls*{asr} errors, and further refine the acoustic support features. We also plan to systematically vary the window length and overlap to quantify the effect of temporal granularity, and to conduct a broader threshold-sensitivity analysis. Regarding temporal modeling, explicit absolute or relative positional encodings could be incorporated to provide the model with direct information about window order. In addition, visualizing the learned \gls*{mil} weights and comparing them with temporal \gls*{ah} annotations would help determine whether the internal weighting aligns with meaningful \gls*{ah} episodes. Finally, facial expressions, head motion, body gestures, and other visual cues could be incorporated as a complementary stream to quantify their incremental contribution over the proposed frame-independent audio--text baseline.

\section*{Acknowledgments}
The authors gratefully acknowledge the \textit{Pontifícia Universidade Católica do Paraná}~(PUCPR) for the financial support that enabled participation in the conference. The authors also acknowledge support from the \textit{Conselho Nacional de Desenvolvimento Científico e Tecnológico}~(CNPq) under Grants 406030/2023-5 and 306878/2022-4, and from the \textit{Fundação Araucária}, in partnership with the \textit{Secretaria da Ciência, Tecnologia e Ensino Superior do Estado do Paraná}~(SETI-PR), under Grant No. 653/2025~(FA/UNIVERSAL).

\bibliographystyle{splncs04}
\bibliography{references}
\end{document}